\documentclass[letterpaper, 10 pt, conference]{ieeeconf}  

\IEEEoverridecommandlockouts                              

\usepackage{graphicx}
\usepackage{graphics} 
\usepackage{amsmath} 
\usepackage{amssymb}  
\usepackage{tabularx}
\usepackage{multirow}
\usepackage{gensymb} 
\usepackage{algorithm} 
\usepackage{algpseudocode} 

\newcommand{\mb}[1]{\mathbf{#1}}

\title{\LARGE \bf
Reconstructing Vechicles from a Single Image: Shape Priors for Road Scene Understanding
}

\author{J. Krishna Murthy$^{1}$, G.V. Sai Krishna$^{1}$, Falak Chhaya$^{1}$, and K. Madhava Krishna$^{1}$
\thanks{$^{1}$J. Krishna Murthy, G. Sai Krishna, Falak Chhaya, and K. Madhava Krishna are with Robotics Research Center, International Institute of Information Technology, Hyderabad, India.
        {\tt\small krrish94@gmail.com mkrishna@iiit.ac.in}}%
}

\begin{document}

\maketitle
\thispagestyle{empty}
\pagestyle{empty}

\begin{abstract}

We present an approach for reconstructing vehicles from a single (RGB) image, in the context of autonomous driving. Though the problem appears to be ill-posed, we demonstrate that prior knowledge about how 3D shapes of vehicles project to an image can be used to reason about the reverse process, i.e., how shapes (back-)project from 2D to 3D. We encode this knowledge in \emph{shape priors}, which are learnt over a small keypoint-annotated dataset. We then formulate a shape-aware adjustment problem that uses the learnt shape priors to recover the 3D pose and shape of a query object from an image. For shape representation and inference, we leverage recent successes of Convolutional Neural Networks (CNNs) for the task of object and keypoint localization, and train a novel cascaded fully-convolutional architecture to localize vehicle \emph{keypoints} in images. The shape-aware adjustment then robustly recovers shape (3D locations of the detected keypoints) while simultaneously filling in occluded keypoints. To tackle estimation errors incurred due to erroneously detected keypoints, we use an Iteratively Re-weighted Least Squares (IRLS) scheme for robust optimization, and as a by-product characterize noise models for each predicted keypoint. We evaluate our approach on autonomous driving benchmarks, and present superior results to existing monocular, as well as stereo approaches.

\end{abstract}

\section{INTRODUCTION}

With the advent of autonomous driving, the robot vision community has been devoting siginficant attention to understanding road scenes. Many of the successful approaches to road scene understanding make extensive use of LiDAR or stereo camera rigs. However, there has been increasing interest in replacing these expensive systems with cheap off-the-shelf cameras. This poses many interesting challenges, which are primary motivating factors for the current paper.

We focus on the specific problem of recovering 3D shape and pose of vehicles, given a single (RGB) image. Our approach is based on the premise that humans are able to perceive 3D structure from a single image, owing to their vast prior knowledge on how various 3D shapes project to 2D. Using this prior knowledge, they perform efficient inference of the reverse process, i.e., the 3D structure from 2D appearance. We attempt to endow machines with the same capability, by learning \emph{shape priors}, and using them along with other constraints arising from projective imaging geometry in a robust optimization framework.

Ability to accurately detect objects and their corresponding part locations (\emph{keypoints}) has been shown to benefit pose and shape estimation \cite{zia_IJCV_2015}. To this end, we leverage recent successes of Convolutional Neural Networks (CNNs) in related visual perception tasks \cite{VpsKps, facial_landmark, flowing_convnets}, and train fully-convolutional regressors for extracting keypoints from a detected object. We show that a small keypoint-annotated training set suffices to train keypoint regressors, and in the subsequent learning of shape priors.

\begin{figure}[t]
\centering
\includegraphics[scale=0.27]{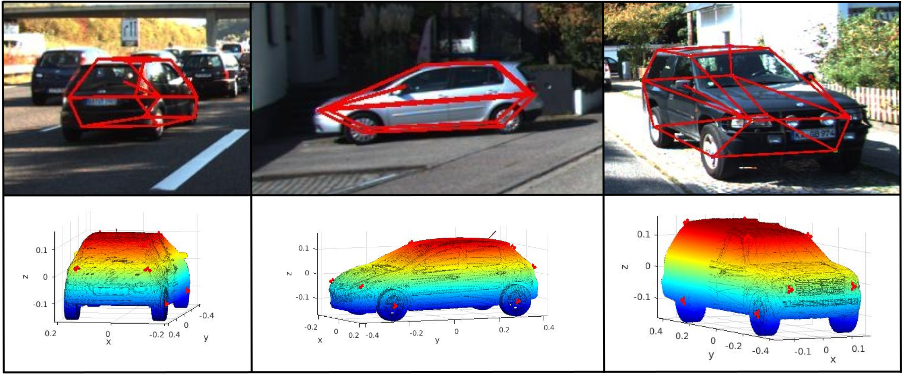}
\caption{Examples of 3D shapes estimated by the proposed pipeline for instances with varying shape and pose. Just using a single image, the algorithm extracts the locations of keypoints in 3D. The algorithm also extracts 3D (and subsequent 2D) locations for occluded keypoints. \textbf{Top:} Estimated 3D shapes projected down to the image. \textbf{Bottom}: Estimated pose and shape shown by rendering the \emph{closest} CAD model to the query instance.}
\label{teaser}
\vspace{-0.5cm}
\end{figure}

\subsubsection*{\textbf{Contributions}}
The primary contribution of this paper is a novel \emph{shape-aware adjustment} scheme which estimates the 3D pose and shape of a vehicle, given the shape prior and keypoint detections. The proposed formulation works even when some keypoints are not detected (due to occlusion) and robustly fills in the missing keypoints. Moreover, it accounts for imperfect keypoint localization and recovers noise models for each detected keypoint by using an Iteratively Reweighted Least Squares (IRLS) scheme. We quantitatively demonstrate that such a reweighting scheme can applied to standard pose estimation pipelines such as \cite{ASPnP} and it boosts accuracy by more than an astounding $90\%$!

\begin{figure*}[t]
\centering
\includegraphics[scale = 0.45]{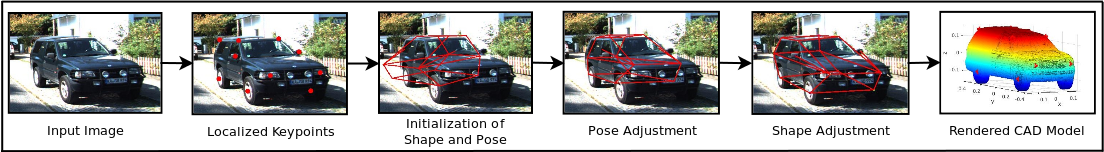}
\caption{Illustration of the proposed pipeline. \textbf{Left to Right:} The input image is passed through the CNN-cascade to accurately localize keypoints. The initialization is (usually) with an incorrect pose and shape. Pose adjustment aligns the initial wireframe with the actual shape. Shape adjustment optimizes for the 3D shape that best explains the 2D keypoint evidence. Note how the headlights and wheels do not project correctly after pose adjustment, but are refined by shape adjustment. We also render the \emph{closest} CAD model (under the Hausdorff distance metric) to the estimated wireframe.}
\label{pipeline}
\vspace{-0.5cm}
\end{figure*}

Another contribution of the paper is a novel architecture for keypoint localization which we call a \emph{CNN-cascade}. CNN-cascade performs the task of accurate keypoint localization and aids the 3D pose and shape adjustment. To guide future efforts towards keypoint localization, we release a dataset comprising of keypoint annotations for a subset of vehicles from the KITTI \cite{KITTI} autonomous driving benchmark.

\subsubsection*{\textbf{Central Idea}}
Estimating shape and pose simultaneously from a single image is an ill-posed problem, and suffers from several ambiguities as demonstrated in \cite{symmetry_NRSfM}. Most notably, when the object pose is unconstrained, optimizing on shape reprojection error results in arbitrary deformations in the estimated 3D shape. The central idea of this paper is to \emph{decouple the pose and shape estimation problems}. Our approach can be viewed as a coarse-to-fine adjustment, where we first obtain the coarse shape of the object by aligning the mean shape for the object category with the query instance. Then, we capture keypoint evidence which is specific to the instance, and run a novel shape-adjustment procedure which ensures that the optimized shape satisfies geometric constraints for the object category (such as planarity, symmetry, etc.), and does not deviate much from the shape prior for the object category. A set of sample outputs from our approach is shown in Fig. \ref{teaser}. The pipeline is summarized in Fig. \ref{pipeline}.

\subsubsection*{\textbf{Evaluation}}
We perform an extensive analysis of the proposed approach on the KITTI \cite{KITTI} benchmark for autonomous driving. We evaluate our approach on more than $14,000$ vehicles and demonstrate superior performance with respect to published monocular and stereo leaders by upto $27\%$ in terms of pose accuracy. \footnote{The code and dataset used in this work will be made publicly available.}



\section{RELATED WORK}

The last decade-and-a-half has witnessed a huge volume of work on high-level perception for autonomous driving. Many successful approaches make use of information from multiple cues, most notably Stereo and LIDAR. In this section we briefly review work pertaining to monocular perception, while discussing the differences in the underlying assumptions/approach.

Philosophically, the closest approach to ours is the one by Zia, et al. \cite{zia_IJCV_2015}, where 3D shape representations for vehicles were learnt from annotated CAD models. Vehicle shape was represented as an ordered collection of 3D part locations and random forest classifiers were employed for localizing 2D parts given an object bounding box. The approach was extended to handle information from multiple views in \cite{falak_ICRA}. However, the shape inference mechanism used in \cite{zia_IJCV_2015,falak_ICRA} was stochastic hill climbing, owing to the highly non-convex and non-smooth cost function, thereby resulting in a very slow, iterative optimization procedure.  Moreover, part localization was performed using random forests, which was slow too. In contrast, our optimization formulation comprises of smooth functions and can be solved using non-linear least squares, and our part localization pipeline is fast.

In a sequence of works, \cite{chandraker2014, chandraker2015} have developed a real-time monocular SfM system for autonomous driving. However, a vehicle is represented as a 3D bounding box; no shape information is involved. Further, information from multiple frames is used. Our approach attempts to use information from only a single image, as opposed to a video sequence.

In \cite{tulsiani_PAMI}, Non-Rigid SfM \cite{NRSfM} is used to learn shape representations over a keypoint-annotated dataset. Using the learnt \emph{average shape} representations and \emph{deformation modes}, dense point clouds of vehicles are reconstructed from a single image. We differ from \cite{tulsiani_PAMI} in that we capture inherent geometric constraints that vehicles exhibit (symmetry and planarity, for instance) which would result in more meaningful shape estimates. This was partially addressed in \cite{symmetry_NRSfM}, which tries to use symmetry and Manhattan properties of objects to recover more meaningful shapes. However, the reconstructions assume a weak-orthographic projection model, whereas we use a projective camera model and a globally optimal pose estimation pipeline.

All the above methods rely on some form of object part detection to produce meaningful reconstructions. For instance, Zia, et. al. \cite{zia_IJCV_2015} rely on Deformable Part Models (DPMs) and random forests. With the promise demonstrated by Convolutional Neural Networks (CNNs) in object detection tasks, they have also been used for the task of viewpoint and keypoint prediction in \cite{VpsKps}. Apart from \cite{VpsKps}, convolutional architectures for keypoint localization have been proposed in \cite{facial_landmark, flowing_convnets}. Motivated by the promise shown by CNNs, we train a cascaded architecture of fully-convolutional networks for keypoint localization.

Current published monocular and stereo competitors for 3D object detection are \cite{mono3d} and \cite{objprop3d} respectively. Both these approaches operate at the level of estimating 3D oriented bounding boxes for car detection. We comprehensively analyze the performance of the proposed approach and demonstrate a significant performance boost over \cite{mono3d,objprop3d}. Note that we also learn a more detailed 3D representation than 3D bounding boxes, which is desirable for reconstruction.

\section{OUR APPROACH}

Simultaneous pose and shape estimation of vehicles is ill-posed when only a single image is available \cite{symmetry_NRSfM}. Guided by the motivation that humans make use of prior information about the vehicle to reason about 3D shape and pose, we decouple the pose and shape estimation problems. Fig. \ref{pipeline} illustrates the overall picture of the proposed pipeline.

\subsection{Shape Priors}

Our approach uses \emph{shape priors} to model the 3D representation of an object category. More specifically, a shape prior is an ordered collection of 3D vertices (\emph{keypoints}, or \emph{parts}) for the \emph{average shape} of an instance from the object category. By \emph{keypoints}, we refer to various semantic parts of an object category that are common to all instances of that category. For example, in the case of cars, potential candidates for keypoints are wheel centers, headlights, rooftop corners, etc. Previous approaches \cite{zia_IJCV_2015,falak_ICRA} made use of a large database of keypoint-annotated CAD models to define shape priors. On the other hand, we make use of a small 2D keypoint-annotated image set consisting of nearly 300 instances from the PASCAL3D+ \cite{PASCAL3D} dataset.

\subsubsection*{\textbf{Learning Shape Priors from 2D Data}}

Given a keypoint-annotated dataset, the conventional approach to reconstructing 3D keypoints is to run a Structure-from-Motion (SfM) pipeline using keypoints as correspondences. However, this is not a suitable approach since keypoint correspondences give a valid reconstruction only when the \emph{same rigid} instance is observed across multiple images. We would like to have as many \emph{different} instances as possible, to capture a large part of the intra-class shape variations. But, when we have different instances, it is no longer possible to use keypoints as correspondeces. To overcome this difficulty, we use the EM-PPCA method of \cite{tulsiani_PAMI,NRSfM} and cast the problem of \emph{lifting the dataset from 2D to 3D} into the framework of Non-Rigid Structure-from-Motion (NRSfM). Consider that we are given $M$ instances each annotated with $K$ keypoints (2D shapes), where $\mb{s}_{i,m}$ ($i \in \{1..K\}, m \in \{1..M\} $) represents the $i^{th}$ keypoint of instance $m$. As in \cite{tulsiani_PAMI}, we assume that the projection model is weakly-orthographic (note that this assumption is made only until we learn shape priors; we relax this soon). We denote the 3D shape of an instance $m$ by $\mb{S}_m$, i.e., $\mb{S}_m = \left[\mb{s}_{1,m}^T, ..., \mb{s}_{K,m}^T\right]^T$ and the (orthographic $2 \times 3$) rotation and translation of the camera for the instance by $\mb{R}_m$ and $\mb{t}_m$. NRSfM hypothesizes that real-world shapes are confined to a low-dimensional \emph{basis} of the entire shape space. We express a specific shape instance as the sum of the mean shape of the object category $\bar{\mb{S}}$ deformed using a linear combination of vectors from a set of $N$ \emph{basis shapes} $\mb{V} = \left[\mb{V}_1, ..., \mb{V}_N\right]$. Then, the NRSfM problem \cite{NRSfM} involves maximizing the likelihood of the following model.
\begin{equation}
\begin{split}
\begin{aligned}[c]
&\mb{s}_{i,m} = c_m\mb{R}_m (\mb{S}_{i,m} + \mb{t}_m) + \mb{\delta}_{i,m} \\
&\mb{S}_m = \bar{\mb{S}} + \sum_{j=1}^{N} \lambda_{j,m}\mb{V}_j \\
& \mb{\delta}_{i,m} \sim \mathcal{N}(\mb{0},\sigma^2\mb{I}) \; \; \; \; \; \lambda_{j,m} \sim \mathcal{N}(0,1)
& \mb{R}_i^T\mb{R}_i = \mb{I}
\end{aligned}
\end{split}
\label{eqn:NRSfM}
\end{equation}
In the above model, $\lambda_{j,m}$ are the scalar weights (latent variables) of the shape combination, and $\mb{\delta}_{i,f}$ models observation noise. $c_m$ is the weak-orthographic scaling factor.

Note that all instances would have some missing keypoints, due to occlusion. Hence, the likelihood of the model is maximized using the EM algorithm. Missing data is filled in the E-step using the method of \cite{NRSfM}. We then have a mean shape and basis vectors that characterize various deformation modes of the mean shape. For the case of a car, choosing the first five basis vectors captured more than $99.99\%$ of the shape variations.

\subsection{Pose Adjustment}

For correctly estimating the 3D pose of an object, we propose a variant of Perspective n-Point (PnP) \cite{ASPnP} to obtain a robust, globally optimal solution. The pose adjustment pipeline operates in two phases, viz. keypoint localization and robust alignment.

\subsubsection*{\textbf{Keypoint localization}}
\begin{figure}[h]
\centering
\includegraphics[scale=0.35]{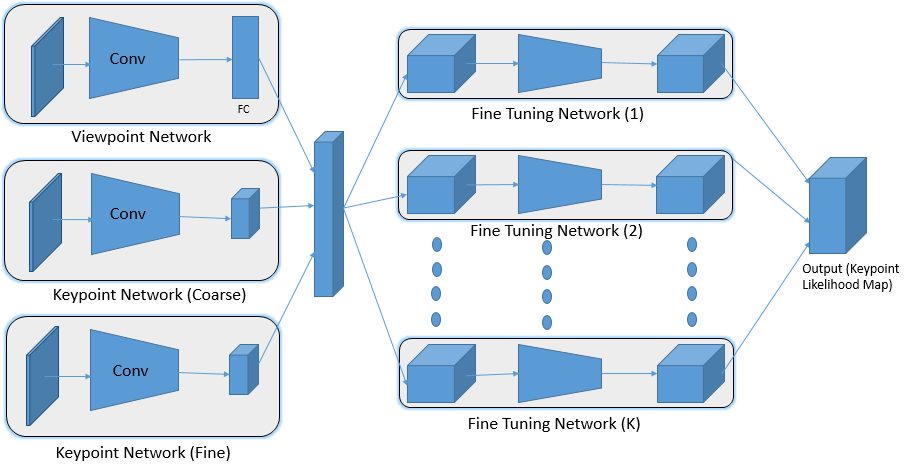}
\caption{Architecture of our Keypoint Localization Network}
\label{fig:CNN}
\end{figure}

The approach relies on localization of the keypoints for the object class in the image. Leveraging the recent successes of Convolutional Neural Networks for the task of keypoint localization \cite{VpsKps,facial_landmark,flowing_convnets}, we train a fully convolutional regressor to accurately localize keypoints given a detected object instance as input. Rather than relying entirely on monolithic networks that operate at the instance level as in \cite{VpsKps,flowing_convnets}, we propose augmenting such networks with a cascaded architecture consisting of small, keypoint-specific subnetworks, following the broad idea of \cite{facial_landmark}. Specifically, we use the keypoints detected by \cite{VpsKps} (denoted KNet) and further refine them using keypoint-specific finetuning networks to reduce mis-predictions and to obtain a more precise localization. Around every keypoint location output by KNet, we sample $N$ random patches of size $32 \times 32$, which is input to the finetuning network for the corresponding keypoint. The finetuning networks are trained in such a way that they output a non-parametric probability distribution of keypoint likelihoods over the input patch, rather than regressing to a vector denoting keypoint coordinates, as in \cite{facial_landmark}. We aggregate and normalize keypoint likelihoods over the $N$ patches, and choose the location of the maxima of the normalized keypoint likelihood map to be the estimated location of the keypoint. The architecture of our network is shown in Fig. \ref{fig:CNN}.

\subsubsection*{\textbf{Robust Alignment}}
Given keypoint locations from the CNN-cascade in the earlier stage, we now formulate a robust PnP problem and solve it to obtain a globally optimal solution using a Gr\"{o}ebner basis solver.
The usual PnP problem for estimating the camera extrinsics ($\mb{R},\mb{t}$), given a set of $n$ 3D points $\mb{X}_i$ and their 2D correspondences $\mb{x}_i$ can be stated as the following optimization problem (Here, $\mb{K}$ is the \emph{projective} camera intrinsic matrix).
\begin{equation}
\begin{aligned}[c]
\underset{\mb{R},\mb{t}}{\text{min}} & \sum_{i=1}^{n} \| \mb{K} (\mb{R}\mb{X}_i+\mb{t}) - \mb{x}_i \|^2 \\
& \text{s.t.} \; \; \; \; \mb{R}^T\mb{R} = \mb{I}
\end{aligned}
\end{equation}

It is hard to achieve a global optima for the above problem, owing to the rotation matrix constraint. However, \cite{ASPnP} formulates a polynomial cost function by parametrizing the rotation matrix as a non-unit quaternion and uses a Gr\"{o}ebner basis solver to obtain a global optimum. It can briefly be summarized as
\begin{equation}
\centering
\begin{aligned}[c]
& \underset{a,b,c,d}{\text{min}} \| \mb{M\alpha} \|^2
\end{aligned}
\end{equation}
where $\mb{\alpha}$ is the Gr\"{o}ebner basis containing the parameters of the quaternion, namely $a,b,c,d$ and the translation, and $M$ is the matrix that holds the coefficients of the Gr\"{o}ebner basis.

This method has one major drawback in that it is intolerant to keypoint localization errors. We extend the formulation in \cite{ASPnP} to provide a more robust solution in presence of keypoint localization errors. Specifically, we use the method of Iteratively Re-weighted Least Squares (IRLS) to weigh each observation (keypoint) in the cost function and solve it iteratively, updating the weights of each observation. To assign weights to a particular observation, we take into account two confidence measures. First, we consider the keypoint likelihood as output from the CNN-cascade $w_{cnn}$. Second, we consider the prior probability of the keypoint being visible from the pose estimate at time $t$ $w_{vis}(t)$, computed by ray-tracing. The weight assigned to the $i^{th}$ keypoint at time $t$ is
\begin{equation}
\centering
w_i(t) = \mu_0 w_{cnn} + (1-\mu_0) w_{vis}(t)
\label{eqn:IRLS_weight_init}
\end{equation}
In each iteration, we solve the PnP problem to obtain estimates for ($\mb{R},\mb{t}$). We update the weights for each observation in the following manner.
\begin{equation}
\centering
w_i(t+1) = \mu_1 w_i(t) + (1-\mu_1) \left[ \mu_2 e_i(t) + (1-\mu_2) w_{vis}(t) \right]
\label{eqn:IRLS_weight_update}
\end{equation}
Here, $e_i(t)$ is the reprojection error in the $i^{th}$ keypoint, normalized over all instances. $\mu_0$, $\mu_1$ and $\mu_2$ are hyperparameters that determine the importance of each term. The final optimization problem can be written in the form shown in Eq. \ref{eqn:aspnp_irls}, where $\mb{W}$ is a diagonal matrix of weights. Algorithm \ref{algo:ASPnP_IRLS} presents an overview of the process.

\begin{equation}
\centering
\begin{aligned}[c]
& \underset{a,b,c,d}{\text{min}} \| \mb{MW\alpha} \|^2
\end{aligned}
\label{eqn:aspnp_irls}
\end{equation}

\begin{algorithm}[H]
\caption{ASPnP + IRLS}
\label{algo:ASPnP_IRLS}
\begin{algorithmic}[1]
\State Initialize weights as in Eq. \ref{eqn:IRLS_weight_init}
\State Initialze ($\mb{R,t}$) by solving the PnP problem in Eq. \ref{eqn:aspnp_irls}
\State $t \gets 0$
\While{$t \le $ MAX\_ITERS (usually set to $5$)}
	\State Compute reprojection error err$_i(t)$ for each keypoint
	\State Normalize all reprojection errors
	\State Update weights as in Eq. \ref{eqn:IRLS_weight_update}
	\State Solve for ($\mb{R,t}$) by solving Eq. \ref{eqn:aspnp_irls}
	\State $t \gets t + 1$
\EndWhile
\end{algorithmic}
\end{algorithm}

Note that this variant of IRLS converges in just 4-5 iterations in all cases. We use the final set of weights obtained from IRLS as an indicator of how confident (or noisy) the keypoint is.

\subsection{Shape-Aware Adjustment}

Using the pose estimated from the previous stage as an initialization, we now formulate a shape-aware adjustment problem that captures---in addition to standard reprojection error---constraints generated from planarity and symmetry in the object category.

Recall that in Eq. \ref{eqn:NRSfM} we defined the shape of an instance to be expressible in terms of the sum of the mean shape and a linear combination of the basis vectors. If $\mb{X}_i$ denotes the $i^{th}$ 3D keypoint location, it can be expressed in terms of the mean location for $\mb{X}_i$ (denoted $\bar{\mb{X}}_i$), and its deformation basis and coefficients $\mb{V}_i = \left[\mb{V}_{i,1}, ..., \mb{V}_{i,N}\right]$ and $\mb{\lambda}_{j}$ respectively. Note that the deformation basis $\mb{V}_i$ is specific to each keypoint (obtained by extracting specific rows from the $\mb{V}$ matrix resulting from NRSfM), but the coefficients $\mb{\lambda}_j$ are common to all keypoints. If this was not the case, the resultant deformations of varying $\mb{\lambda}_j$ would be unconstrained.
\begin{equation}
\mb{X}_i = \bar{\mb{X}}_i + \sum_{j=1}^{N} \lambda_{i,j}\mb{V}_{i,j}
\end{equation}

\subsubsection*{\textbf{Reprojection Error}}
We define the reprojection error of a set of points formally as

\begin{equation}
E_{reproj} = \sum_{i=1}^{K} \| \mb{K}\mb{X}_i - \mb{x}_i \|^2
\end{equation}

\subsubsection*{\textbf{Planarization}}
We exploit the fact that each vehicle consists of a set of planar (or quasi-planar) surfaces. For instance, the centers of the wheels of a car are planar; so are the corners of the rooftop in most cases. We consider the shape $\mb{S}$ to be a quad-mesh consisting of a few planar faces $\mb{F}$. For each face $f \in \mb{F}$, we define four variables ($\mb{n}_f,d_f$), where $\mb{n}_f$ is a $3$-vector representing the normal of the plane, and $d_f$ is a scalar that represents the distance of the plane from the origin. We solve for these variables by defining a planarization energy as
\begin{equation}
E_{planar} = \sum_{f \in \mb{F}} \left( \sum_{v \in f}  \| \mb{n}_f^T\mb{X}_v + d_f \|^2 \right) - \mu_f \left(1-\mb{n}^T_f\mb{n}_f\right)
\end{equation}
The first term encourages all four points on the face (four points, since we assume a quad mesh) to be planar, while the second term encodes the constraint that the normals must be of unit magnitude. Wherever applicable, we also constrain the normals so that they are parallel to the ground plane normal. Furthermore, it is not enough if certain points are planar; they must also be rectangular. These requirements are imposed as further constraints.

\subsubsection*{\textbf{Symmetrization}} All real-world cars exhibit symmetry about their medial plane. However, we find this fact rarely exploited for monocular reconstruction. Symmetry constraints are imposed for reconstruction in \cite{symmetry_NRSfM}, but they propose a factorization-based solution for the weak-orthographic projection case. On the other hand, we impose symmetry constraints for the projective case and solve it using nonlinear least squares. We consider two sets of corresponding symmetric points - $\mb{X}_r \in \mb{R}$ to the right of the medial plane, and correspondingly $\mb{X}_l \in \mb{L}$ to the left of the medial plane. If $\mb{T} = (\mb{n}_{med}, d_{med})$ is the medial plane, we define the symmetrization energy as prescribed in \cite{symmetrization}.
\begin{equation}
E_{sym} = \sum_{\mb{X}_r \in \mb{R}} \| \mb{X}_r + 2(d_{med}-\mb{n}_{med}^T\mb{X}_r)\mb{n}_{med} - \mb{X}_l \|^2
\end{equation}

\subsubsection*{\textbf{Regularization}}
To ensure that the energies proposed herewith result in realistic output values, we define a set of regularization terms.
We encourage the length $\mathcal{L}(\mb{S})$, width $\mathcal{W}(\mb{S})$, and height $\mathcal{H}(\mb{S})$ of each shape $\mb{S}$ to be \emph{close} to their prior values, computed over the dataset. If $\bar{\mb{l}}$, $\bar{\mb{w}}$, and $\bar{\mb{h}}$ are the prior lengths, widths, and heights respectively, then the regularizers for dimensions are
\begin{equation}
E_{dim} = \mu_l \| \mathcal{L}(\mb{S}) - \bar{\mb{l}} \|^2 + \mu_w \| \mathcal{W}(\mb{S}) - \bar{\mb{w}} \|^2 + \mu_h \| \mathcal{H}(\mb{S}) - \bar{\mb{h}} \|^2
\end{equation}

Further, we use a modification of the well-known Laplacian smoothing regularizer which encourages each vertex to remain close to the centroid of its neighbors. We have found it to perform slightly better than the usual Euclidean distance regularizer.
\begin{equation}
E_{lap} = \sum_{i=1}^{K} \| \mb{X}_i - \left( \sum_{j \in \text{Nbd}(\mb{X_i})} e^{-\|\mb{X}_i-\mb{X}_j)\|^2} \mb{X}_j \right) \|^2
\end{equation}

\subsubsection*{\textbf{Initialization}}
To initialize our pose adjustment pipeline, we use 3D bounding boxes from \cite{mono3d}. We rotate and translate the mean shape to the center of the estimated bounding box, and scale the wireframe such that it fits the box. Although such an initialization is not mandatory (the ASPnP + IRLS is fairly robust to initialization errors (see Fig. \ref{pipeline})), we find that this results in a slight performance gain.

\subsubsection*{\textbf{Shape-Aware Adjustment}}
Our final shape-aware adjustment is the conglomeration of all the above terms (with appropriate weighing coefficients), and can be written as
\begin{equation}
E_{total} = \eta_1 E_{reproj} + \eta_2 E_{planar} + \eta_3 E_{sym} + \eta_4 E_{dim} + \eta_5 E_{lap}
\end{equation}

All terms in the equations are easily differentiable (unlike those in \cite{zia_IJCV_2015,falak_ICRA}), and can be solved using a standard non-linear least squares solver. Moreover, since we are solving for the shape of one instance, the number of variables involved is small, which results in near real-time performance. Again, we emphasize that, we do not solve for the locations of the 3D points $\mb{X}_i$; we instead solve for the weights $\lambda_{j,m}$ of the basis vector combination (see Eq. \ref{eqn:NRSfM}). We apply 5 iterations of IRLS weight updates, to reduce the effect of incorrect observations on the estimated shape.

\section{RESULTS}

We perform a thorough qualitative and quantitative analysis of the proposed approach on the \emph{Car} class of the challenging KITTI object detection and pose estimation benchmark \cite{KITTI}. Since we decouple the problem of shape optimization from that of pose estimation, we evaluate each of them independently. To indicate that this evaluation is fair, we also conduct experiments to verify that the shape optimization does not result in significant changes in object pose. We also present an analysis of per-keypoint localization errors. Finally, we show qualitative results (Fig. \ref{qualitative_base}) which indicate that the proposed approach works over a wide range of vehicle shapes.

\subsubsection*{\textbf{Dataset}} The dataset consists of 7481 training images, with specified train and validation splits. Based on the  levels of occlusion and truncation, the evaluation has been split into indicated in the \emph{Easy}, \emph{Moderate}, and \emph{Hard} regimes. We train the CNN-cascade by annotating data from the train split. We evaluate our approach against other competitors on the validation split.

To evaluate the pose estimated by the proposed approach, we use all the 3424 images (14327 instances) from the validation split. However, for the evaluation of part detectors, we manually annotate 500 images from the validation split, and evaluate our algorithm on it, akin to \cite{zia_IJCV_2015}.

\subsubsection*{\textbf{Keypoint-annotations for KITTI}} Keypoint localization is increasingly being recognized as a tool to develop finer understanding of objects. To guide further research in this direction, especially in the context of autonomous driving, we annotate keypoints for cars in a subset of the KITTI object dataset. To ensure consistency of the annotated keypoints with their 3D locations, we follow the procedure adopted in creating the PASCAL3D+ dataset \cite{PASCAL3D}. Specifically, we use the 2D keypoint coordinates and initialize a PnP (Perspective n-Point) algorithm that uses reference CAD models to correct incorrectly marked 2D keypoints. We intend to make the annotations publicly available.

\subsubsection*{\textbf{Metrics}} To evaluate the pose estimated by our approach, we use the Average Orientation Precision (AOP) metric proposed in \cite{PASCAL3D}, as well as the mean absolute difference in orientation estimates. For our AOP criterion, the detection output is considered correct if and only if the overlap ratio with the ground-truth bounding box is greater than $70\%$ and the difference between the predicted and ground-truth viewpoints is less than a particular threshold. We characterize the performance of various approaches by evaluating them for various values of the threshold.

To evaluate error in estimated keypoint locations, we use the APK (Average Precision of Keypoints) metric introduced in \cite{APK}. Under this metric, a keypoint is assumed to be accurately localized if it lies within $\alpha*\text{max}(h,w)$, where $h,w$ are the height and width of the bounding box respectively, and we set $\alpha$ to $0.1$, following \cite{VpsKps}.

To evaluate quantitatively the distance between two shapes (meshes), we use the Hausdorff distance metric.

\begin{table*}[!hbt]
\centering
\begin{tabular}{|c||c|c|c||c|c|c||c|c|c|}
\hline
\textbf{Approach} & \multicolumn{9}{c|}{\textbf{Pose Estimation Accuracy} in terms of Average Orientation Precision (\textbf{AOP})} \\
\hline
& \multicolumn{3}{c||}{\emph{Easy}} & \multicolumn{3}{c||}{\emph{Moderate}} & \multicolumn{3}{c|}{\emph{Hard}} \\
\hline
& $\le 5\degree$ & $\le 15\degree$ & $\le 30\degree$ & $\le 5\degree$ & $\le 15\degree$ & $\le 30\degree$ & $\le 5\degree$ & $\le 15\degree$ & $\le 30\degree$ \\
\hline

ObjProp3D \cite{objprop3d} & $40.04$ & $81.37$ & $91.83$ & $36.76$ & $77.41$ & $89.31$ & $33.98$ & $73.22$ & $85.81$ \\

Mono3D \cite{mono3d} & $42.20$ & $84.59$ & $92.90$ & $38.52$ & $80.60$ & $90.08$ & $35.78$ & $76.27$ & $86.67$ \\

ASPnP \cite{ASPnP} & $1.06$ & $6.18$ & $27.26$ & $1.59$ & $6.35$ & $25.40$ & $2.87$ & $6.90$ & $25.29$ \\

\hline

\textbf{Pose Adjustment (Ours)} & $\mb{53.62}$ & $\mb{90.44}$ & $\mb{95.95}$ & $\mb{48.50}$ & $\mb{85.67}$ & $\mb{94.44}$ & $\mb{44.72}$ & $\mb{80.98}$ & $\mb{89.08}$ \\

\hline

\textbf{Percentage Improvement} & $\mb{+27.06\%}$ & $\mb{+6.91}$ & $\mb{+3.28\%}$ & $\mb{+25.91\%}$ & $\mb{+5.41\%}$ & $\mb{+9.92\%}$ & $\mb{+14.34\%}$ & $\mb{+6.17\%}$ & $\mb{+3.77\%}$ \\

\hline

\end{tabular}
\caption{Orientation precision evaluation for various error ranges on the KITTI Object Detection Benchmark. The percentage improvement row is computed with respect the next best performer in each category (column).}
\label{table:poseResultsMain}
\end{table*}

\begin{table*}[!htb]
\centering
\begin{tabular}{|c||c|c|c||c|c|c|}
\hline
\textbf{Approach} & \multicolumn{3}{c||}{\textbf{APK}}& \multicolumn{3}{c|}{\textbf{Hausdorff Distance from Mean Shape}} \\
\hline
& Easy & Moderate & Hard & Easy & Moderate & Hard \\
\hline
Initialization & $47.31$ & $51.65$ & $43.17$ & $0.00$ & $0.00$ & $0.00$ \\
Pose Adjustment & $62.42$ & $62.72$ & $59.89$ & $3.71$ & $4.64$ & $4.79$ \\
Shape Adjustment & $\mb{80.72}$ & $\mb{81.81}$ & $\mb{71.39}$ & $3.75$ & $4.69$ & $4.83$ \\
\hline
\hline
\multicolumn{4}{|c||}{\textbf{Between Pose and Shape Adjustment}} & $\mb{0.27}$ & $\mb{0.28}$ & $\mb{0.04}$ \\
\hline
\end{tabular}
\caption{Effect of each step in the proposed pipeline on shape estimation}
\label{table:shapeResults}
\end{table*}

\subsubsection*{\textbf{Implementation}} The CNNs used for keypoint detection were based on the fully convolutional architecture presented in \cite{VpsKps}, with the only change that the loss function was defined in terms of a non-parametric Gaussian over the true keypoint location. The networks were trained in Caffe \cite{Caffe}, over annotated data from the train split. The provided train/validation split ensured that there was no overlap between images/sequences.

The shape-aware adjustment was implemented using Ceres Solver \cite{Ceres}, and was solved using a dense Schur linear system solver and a Jacobi preconditioner.

%
%

\subsection{Pose Estimation}

To analyze the efficiency of our pose estimation approach , we evaluate it against Mono3D \cite{mono3d}, which is the current monocular competitor for 3D object detection and pose estimation. We also show that, due to inaccuracies in keypoint estimates, the standard ASPnP \cite{ASPnP} formulation struggles to find a suitable rotation and translation for the shape prior, whereas the proposed approach (ASPnP + IRLS) achieves precise pose (to within $5$ degrees) for more than $50\%$ of the samples. Table \ref{table:poseResultsMain} compares our results with that of Mono3D \cite{mono3d} and ObjProp3D \cite{objprop3d}. Note that ObjProp3D is a stereo-based approach and the comparision is thus unfair. On an average, we improve the pose estimation performance with respect to the next-best competitor (for each evaluation category) by $11.42\%$.

We see that, in most cases, we are within $\pm 30\degree$ of the original azimuth, while more than half of the time, we are within $\pm 5\degree$ of the actual azimuth. In Table \ref{table:poseResults2}, we show the mean absolute difference (error) in the pose estimates. Here, we see the robustness that IRLS renders to the pose adjustment pipeline. ASPnP \cite{ASPnP}---though claims global optimiality of its solution---struggles in the presence of noise. It turns out that in all instances of cars, one or more keypoints is self-occluded and hence filled-in only approximately by the keypoint localization module. In our formulation of the IRLS weight update, weights assigned to incorrectly estimated keypoints get reduced every iteration, and have less effect on the final pose estimate.

\begin{table}[!b]
\centering
\begin{tabular}{|c||c|c|c|}
\hline
\textbf{Approach} & \multicolumn{3}{c|}{\textbf{Mean Absolute Error} (in degrees)} \\
\hline
& Easy & Moderate & Hard \\
\hline
ObjProp3D \cite{objprop3d} & $17.32$ & $21.86$ & $26.87$ \\
Mono3D \cite{mono3d} & $15.14$ & $15.15$ & $17.82$ \\
ASPnP \cite{ASPnP} & $98.17$ & $98.82$ & $98.83$ \\
\hline
\textbf{Pose Adjustment (Ours)} & $\mb{8.79}$ & $\mb{12.57}$ & $\mb{16.19}$ \\
\hline
\end{tabular}
\caption{Mean absolute error in pose estimation (deviation from ground-truth pose, in degrees)}
\label{table:poseResults2}
\end{table}

\begin{figure*}[!htbt]
\centering
\includegraphics[scale=0.17]{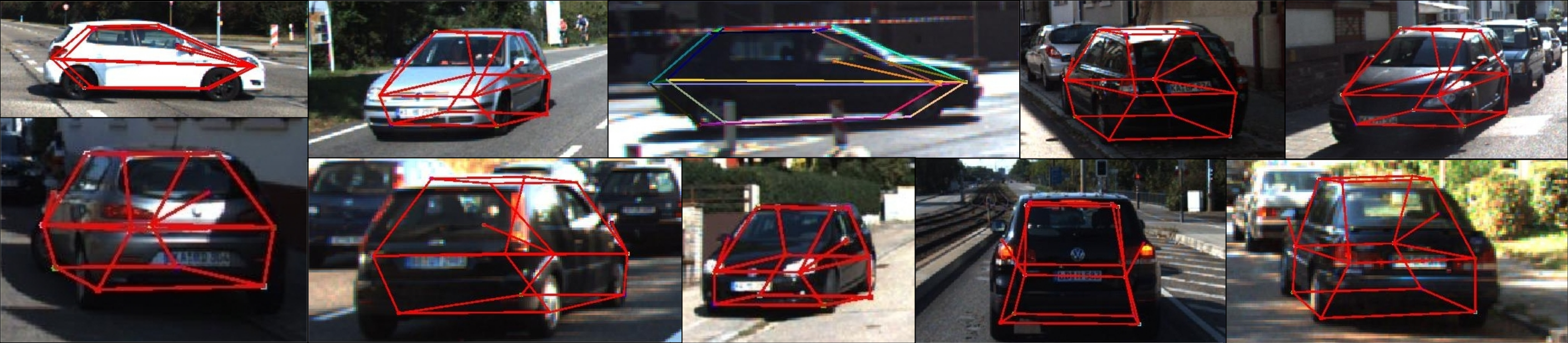}
\caption{Qualitative results showing shape reconstructions from our pipeline.}
\label{qualitative_base}
\end{figure*}

\subsection{Shape-Aware Adjustment}

We now demonstrate the performance of the shape adjuster by quantitifying the precision in part locations before and after shape optimization. Note that reprojection error alone is not a measure of 3D shape correctness, as we could always have an arbitrary shape that could lead to zero reprojection error. To demonstrate that the optimized wireframe is \emph{close} to the shape space of cars, we evaluate a mesh error metric based on the Hausdorff distance of the optimized shape from the shape prior. Further, following \cite{zia_IJCV_2015}, we show qualitative results for 3D shape estimation (Fig. \ref{qualitative_base}). Table \ref{table:shapeResults} shows the perfomance of our shape adjuster on subsets of the easy, moderate, and hard splits. We sampled every 50th car in the easy split, every 75th car in the moderate split, and every 100th car in the hard split (this was done to get a nearly equal number of images in each resultant set). As shown in Table \ref{table:shapeResults}, there is a significant change between the shape prior (initialization) and the output from the pose adjustment module. This is due to the fact that the initialization is erroneously oriented most of the time, and the pose adjustment results in significant rotations being applied to it. Table \ref{table:shapeResults} validates an important claim of this paper, viz. \emph{pose and shape estimation problems can be decoupled}. This claim is supported by the fact that shape adjustment results in small, local changes to the shape model, in comparison with pose adjustment, which results in (possibly) large, global transformations. Due to the scale of the solutions to the pose estimation and the shape estimation problems, it is best that they are solved sequentially.

\begin{figure}[!htb]
\centering
\includegraphics[scale=0.4]{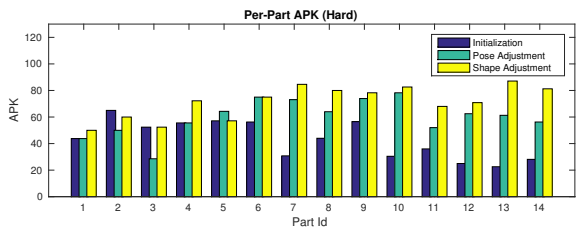}
\caption{Per-Part APK Measure for the \emph{Hard} split. Parts 1-4 correspond to the wheels, 5-6 correspond to the headlights, 7-8 correspond to the taillights, 9-10 correspond to the side-view mirrors, and 11-14 correspond to four corners of the rooftop.}
\label{per_part}
\end{figure}

Fig. \ref{per_part} provides an analysis of APK for each keypoint. This plot strongly justifies the need for shape adjustment. We can observe that, by pose adjustment alone, we improve on APK, but this improvement is strictly \emph{on an average}, i.e., APK averaged over all parts improves, but there are parts (eg. parts 2, 3) where the initialization has a lower APK. After shape adjustment, however, the APK improves significantly for all parts. Note that part 2 (which corresponds to the center of the right front wheel) is one of the least frequently occurring parts in the KITTI object dataset.

\subsection{\textbf{Run-Time Analysis}}

All the pipelines we proposed have been designed with the goal of being applicable in autonomous driving scenarios. Barring the CNN-cascade, all other modules run on a single CPU core. The CNN-cascade runs on an NVIDIA GeForce GTX. Details of running times of various components of the pipeline are furnished in Table \ref{table:runTime}.

\begin{table}[h]
\centering
\begin{tabular}{|c|c|c|}
\hline
\textbf{Module} & \textbf{Runs on} & \textbf{Runtime per instance (in ms)} \\
\hline
Pose Adjustment & CPU & 9.97 ms \\
CNN-cascade & GPU & 200.79 ms \\
Shape Adjustment & CPU & 52 ms \\
\hline
\end{tabular}
\caption{Running times of various components of the pipeline}
\label{table:runTime}
\vspace{-0.5cm}
\end{table}

\section{CONCLUSIONS}

In this work, we presented an approach for estimating the 3D shape and pose of a vehicle from a single image. We decoupled the pose estimation problem from that of shape estimation, and proposed fast, robust algorithms for each of them. We evaluated our approach against published monocular and stereo competitors and demonstrated superior performance, thereby advancing the state-of-the-art. In retrospect, we feel that the bottleneck to achieving near-perfect performance is imprecise keypoint localization, and we identify that as a future research direction. We would also like to explore the benefits of temporal information, wherever available.

\addtolength{\textheight}{-2cm}   



\section*{ACKNOWLEDGMENT}

The authors would like to thank the following people for spending some of their valuable time annotating keypoints on cars from the KITTI object detection benchmark: Abhishek Raj, Sheetal Reddy, Falak Chhaya, Sarthak Sharma, Abhineet Jain, Parv Parkhiya, and Akanksha Baranwal.


\bibliography{refs}
\bibliographystyle{IEEEtran}


\end{document}